\documentclass[letterpaper, 10 pt, conference]{ieeeconf}  

\IEEEoverridecommandlockouts                              

\usepackage{xcolor}
\usepackage{color}

\usepackage{subcaption}
\usepackage{graphicx}
\usepackage[utf8]{inputenc}
\DeclareUnicodeCharacter{2212}{\ensuremath{-}}
\usepackage{url}
\graphicspath{{figures/}}

\usepackage{comment}
\usepackage{amssymb}
\usepackage{amsmath}
\usepackage{booktabs}
\usepackage{dblfloatfix} 
\usepackage{booktabs} 
\usepackage{graphicx}
\usepackage{booktabs}
\usepackage{array}
\usepackage{adjustbox}  
\usepackage{caption} 
\usepackage{xurl}   
\usepackage{hyperref} 
\newcolumntype{C}[1]{>{\centering\arraybackslash}m{#1}}

\title{\LARGE \bf
Semantics-Guided Multimodal Masked Autoencoder Pretraining for 3D BEV Object Detection}

\author{Prabuddhi Wariyapperuma$^{1}$, Rajitha de Silva$^{1}$, Marc Hanheide$^{1}$, Thomas Bohné$^{2}$ and Leonardo Guevara$^{1,*}$%
\thanks{This work was supported by the Engineering and Physical Sciences Research Council and AgriFoRwArdS CDT [EP/S023917/1].}%
\thanks{$^{1}$ University of Lincoln, Lincoln Centre for Autonomous Systems, Lincoln, UK. $^{2}$ University of Cambridge, Institute for Manufacturing, Department of Engineering, Cambridge, UK.}%
\thanks{* Corresponding author: {\tt\small lguevara@lincoln.ac.uk}}}
\begin{document}

\maketitle
\thispagestyle{empty}
\pagestyle{empty}

\begin{abstract}
Accurate 3D bird’s-eye view (BEV) object detection is essential for autonomous driving, and depends strongly on effective multimodal representations from complementary sensors such as cameras and LiDAR. Multimodal masked autoencoders have shown strong potential for learning such representations for downstream 3D BEV object detection. However, existing methods typically apply uniform random masking to camera and LiDAR inputs, treating all regions equally, and learn representations only through masked reconstruction. We propose a semantics-guided multimodal masked autoencoder framework that introduces semantic information during pretraining through two separate components: (i) semantics-guided LiDAR voxel masking, which preserves semantically important LiDAR regions more strongly, and (ii) an auxiliary point-wise LiDAR semantic decoder branch that injects semantic guidance in addition to reconstruction. On BEVFusion 3D object detection, our semantics-guided pretraining strategy improves performance on the nuScenes mini validation set compared to the standard UniM$^2$AE baseline: semantics-guided LiDAR voxel masking yields +1.49\% mean Average Precision (mAP) and +1.66\% nuScenes Detection Score (NDS), while decoder-side point semantic supervision yields +1.39\% mAP and +3.22\% NDS over the baseline.
\end{abstract}

\section{Introduction}
\label{sec:introduction}

Reliable 3D object detection is essential for autonomous driving because planning and safety-critical decision-making depend on accurately localising surrounding actors and obstacles in three-dimensional space. BEV representation has emerged as a strong formulation for this problem because it reduces perspective distortion and provides a unified top-down view of scene layout, making spatial reasoning and sensor fusion more reliable and effective. In particular, multimodal fusion of camera-LiDAR inputs has become central to high-performance 3D BEV object detection, as the two modalities provide complementary semantic and geometric information \cite{bevfusion}.

Building on this, multimodal masked autoencoders have emerged as a promising way to learn transferable camera-LiDAR representations for downstream 3D detection, because masked reconstruction encourages the model to learn meaningful multimodal structure from incomplete inputs. Multimodal Masked Autoencoders with Unified 3D Representation (UniM$^2$AE) follows this paradigm by randomly masking image patches and LiDAR voxels before fusing the two modalities in a unified 3D volume space for reconstruction and downstream detection \cite{unim}. The existing works demonstrate the promise of multimodal masked autoencoders, but they also raise an important question: is uniform random masking together with reconstruction-only learning sufficient for learning the most effective multimodal representations? 

\begin{figure}[t]
\centering
\includegraphics[width=0.682\columnwidth]{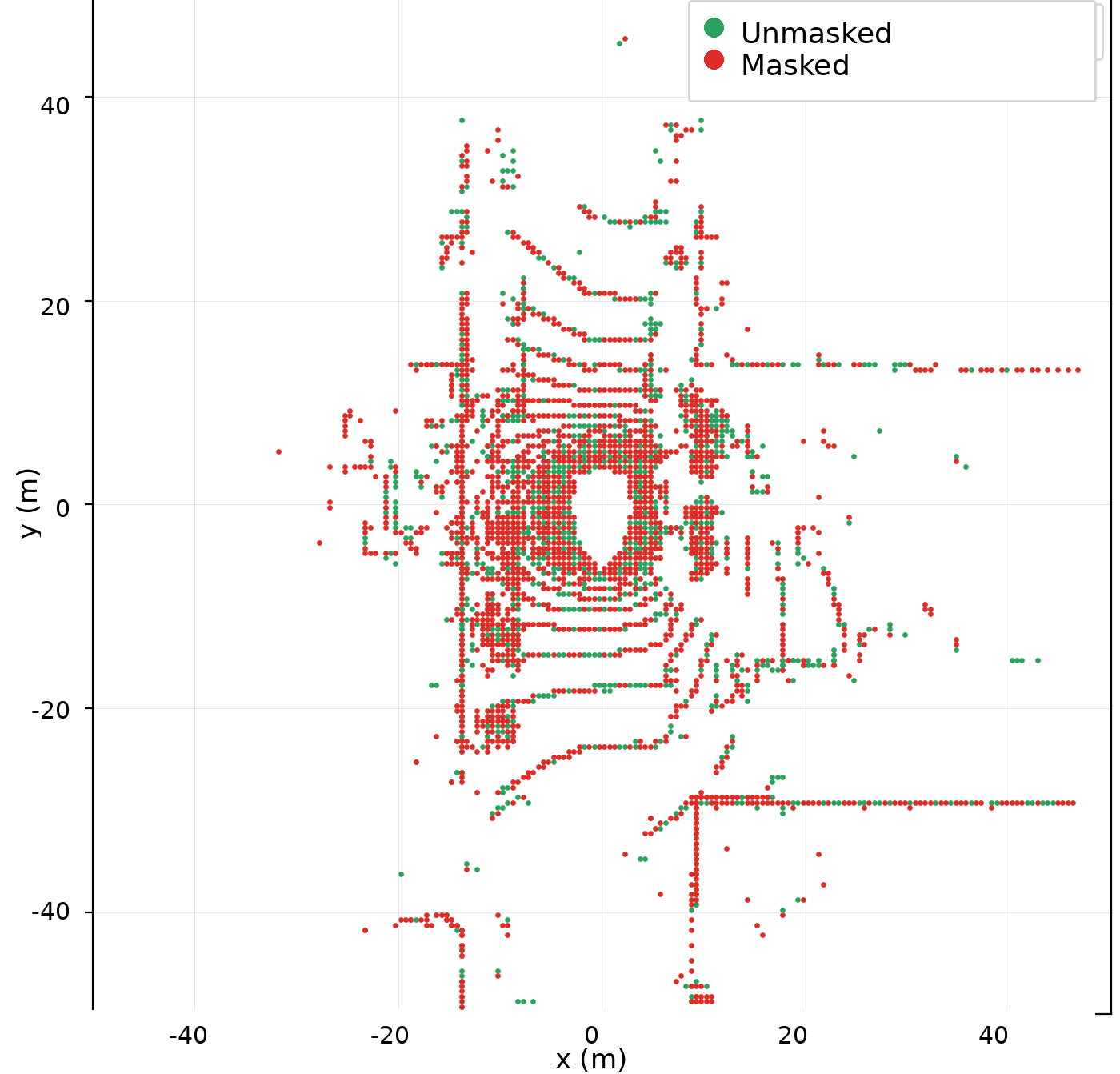}
\caption{Proposed semantics-guided LiDAR voxel masking policy on a sample from the nuScenes mini validation set.
}
\label{fig:priority_masking_vis}
\end{figure}

\begin{figure*}[!t]
\centering
\makebox[\textwidth][c]{\includegraphics[width=2\columnwidth]{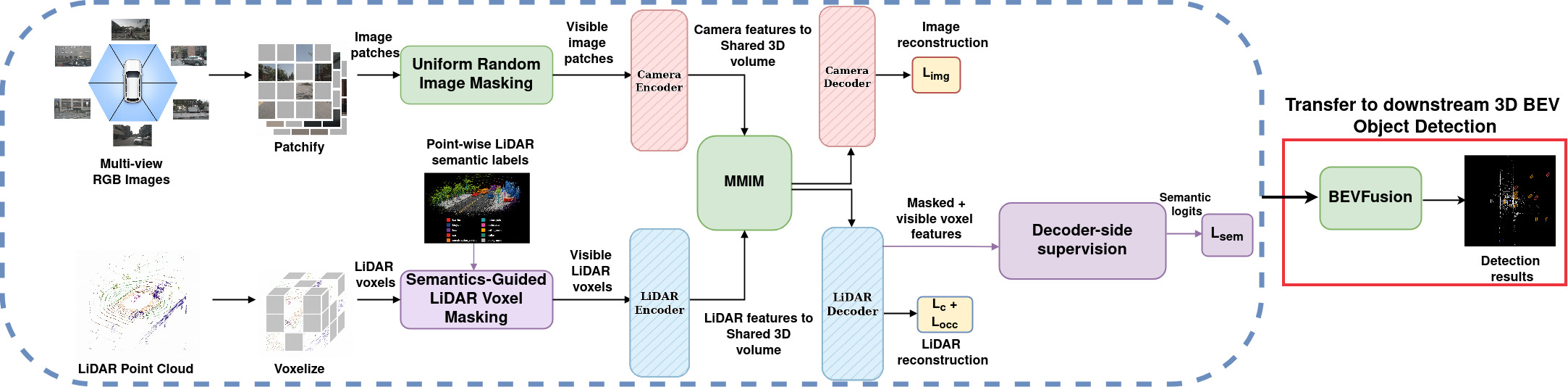}}
\caption{Overview of our semantics-guided multimodal masked autoencoder pretraining framework for downstream 3D BEV object detection. The purple blocks denote the components newly introduced in our method: semantics-guided LiDAR voxel masking and decoder-side point semantic supervision. The dotted blue boundary marks the pretraining-only part of the framework, indicating that all components inside it are used only during pretraining. 
}
\label{fig:method_overview}
\end{figure*}

Existing masked autoencoders largely follow a simple design: randomly hiding part of the input and learning representations by reconstructing the missing content \cite{unim,voxelmae}. More recent work suggests that the masking policy itself can substantially influence what the model learns. In the LiDAR domain, Occupancy-MAE \cite{occupancy} introduces range-aware masking to reflect distance-dependent sparsity, while I2P-MAE \cite{i2pmae} shows that importance-aware masking can preserve semantically important 3D tokens more effectively than uniform random masking. These works highlight the value of preserving semantically important regions during masking. Similar evidence also appears in the image domain, where SemMAE \cite{semmae} demonstrates that semantic-guided masking improves representation learning over uniform random masking. Motivated by these findings, our first contribution introduces semantics-guided masking for LiDAR voxels in multimodal camera-LiDAR masked autoencoders, as illustrated in Fig.~\ref{fig:priority_masking_vis}.


Beyond importance-aware masking, I2P-MAE also introduces semantic targets during pretraining, indicating that semantic guidance can improve masked representation learning beyond pure reconstruction \cite{i2pmae}. Inspired by this, our second contribution introduces auxiliary point-wise LiDAR semantic supervision beyond reconstruction. More broadly, neither semantics-guided masking nor auxiliary semantic supervision has yet been explored in multimodal camera-LiDAR masked autoencoders for 3D BEV detection, where LiDAR voxels are still generally treated as equally maskable units, learning remains dominated by reconstruction, and the impact of semantic information on downstream 3D BEV object detection has not yet been studied.


To address this gap, we propose a semantics-guided multimodal masked autoencoder pretraining framework for downstream 3D BEV object detection. The main contributions of this paper are as follows:
\begin{itemize}
    \item We introduce a semantics-guided LiDAR voxel masking strategy for multimodal masked autoencoder pretraining that preserves semantically important LiDAR regions more strongly than uniform random masking.
    \item We introduce auxiliary point-wise LiDAR semantic supervision during pretraining to complement reconstruction with explicit semantic guidance for LiDAR features.
    \item We show that introducing semantic information during multimodal masked autoencoder pretraining improves downstream 3D BEV object detection on the nuScenes mini validation set compared with the standard UniM$^2$AE baseline.
\end{itemize}



\section{Method}
\label{sec:method}

Our approach introduces semantic guidance to multimodal masked autoencoder pretraining with semantic guided masking and auxiliary semantic supervision. Figure~\ref{fig:method_overview} shows an overview of the proposed framework.

\subsection{Baseline Multimodal Masked Autoencoder Pretraining}
\label{subsec:baseline}

We adopted UniM$^2$AE ~\cite{unim} as the baseline, since it is a strong camera-LiDAR masked autoencoder. 
The LiDAR branch first voxelises the input point cloud into LiDAR voxel tokens, while the camera branch divides the multi-view images into non-overlapping image patch tokens. Uniform random masking is then applied independently to the LiDAR voxels and image patch tokens, and only the resulting visible tokens are processed by the LiDAR and camera encoders, respectively. The visible camera and LiDAR features are projected into a shared 3D volume, fused through the learnable Multimodal 3D Interaction Module (MMIM), and then mapped back to modality-specific decoder inputs. For masked positions, learned mask-token embeddings are inserted at the corresponding voxel or patch locations before decoding to represent the missing content. In both branches, the decoder receives the encoded visible features together with learned mask tokens at masked positions to reconstruct the original camera and LiDAR inputs. The baseline pretraining objective combines masked image and LiDAR voxel reconstruction losses:
\begin{equation}
\label{eq:base_loss}
\mathcal{L}_{\mathrm{base}} = \mathcal{L}_{\mathrm{img}} + \mathcal{L}_{c} + \mathcal{L}_{\mathrm{occ}}
\end{equation}
where $\mathcal{L}_{\mathrm{img}}$ is the mean squared error for masked image reconstruction, $\mathcal{L}_{c}$ is the Chamfer distance loss for LiDAR point-set reconstruction, and $\mathcal{L}_{\mathrm{occ}}$ is the occupancy loss for predicting whether a voxel is empty or occupied.

After pretraining, the decoders are discarded and the pretrained encoders and fusion components are transferred to BEVFusion~\cite{bevfusion} for downstream 3D BEV object detection. In this work, we keep this downstream fine-tuning stage unchanged and introduce semantic information only during pretraining through (i) semantics-guided LiDAR voxel masking and (ii) auxiliary point-wise LiDAR semantic supervision. 

\begin{table*}[t]
\centering
\caption{Downstream 3D BEV object detection results on the nuScenes mini validation set.}
\label{tab:main_results}
\footnotesize
\setlength{\tabcolsep}{4pt}
\begin{tabular}{|l|c|c|c|c|c|c|c|}
\hline
\textbf{Method} & \textbf{mAP (\%) $\uparrow$} & \textbf{NDS (\%) $\uparrow$} & \textbf{mATE $\downarrow$} & \textbf{mASE $\downarrow$} & \textbf{mAOE $\downarrow$} & \textbf{mAVE $\downarrow$} & \textbf{mAAE $\downarrow$} \\
\hline
Baseline (Uniform Random Masking) & 24.72 & 31.41 & 0.5682 & 0.5373 & 1.2709 & 0.6455 & 0.3436 \\
\hline
Semantics-Guided LiDAR Voxel Masking & \textbf{26.21} & 33.07 & 0.5054 & 0.5089 & 1.1032 & 0.6555 & 0.3342 \\
\hline
Decoder-Side Point Semantic Supervision & 26.11 & \textbf{34.63} & \textbf{0.4802} & \textbf{0.5033} & 1.1235 & \textbf{0.5614} & \textbf{0.2976} \\
\hline
Post-MMIM Point Semantic Supervision (Ablation) & 26.09 & 32.96 & 0.4821 & 0.5203 & \textbf{1.0509} & 0.6845 & 0.3217 \\
\hline
\end{tabular}
\end{table*}

\subsection{Semantics-Guided LiDAR Voxel Masking}
\label{subsec:semantic_masking}


The original UniM$^2$AE applies uniform random masking to LiDAR voxels, treating all voxels equally. In this work, we investigated whether different semantic classes affect masked LiDAR reconstruction differently, and whether this influences downstream 3D BEV object detection. To do this, we used LiDAR semantic labels to guide voxel masking during pretraining.


\paragraph{Semantic Class Importance Analysis}
We analysed the importance of each semantic class for masked LiDAR reconstruction. For a target semantic class $c$, we define the set of target voxels as $\mathcal{V}^{(c)} = \{v \mid n_v^{(c)} \ge \tau_c\}$, where $n_v^{(c)}$ denotes the number of points of class $c$ in voxel $v$, and $\tau_c$ is a class-specific threshold. A voxel was therefore assigned to $\mathcal{V}^{(c)}$ if it contained at least $\tau_c$ points from class $c$. To ensure a fair comparison with the baseline, we kept the overall LiDAR masking ratio fixed at the baseline value $\rho$, so that only the semantic composition of the masked voxels changed. We first masked all voxels in $\mathcal{V}^{(c)}$ and then masked the remaining occupied voxels at random until the final LiDAR masking ratio matched the baseline masking ratio $\rho$.

We then evaluated each class-specific masking setting using voxel-level LiDAR reconstruction metrics in Eq.~\eqref{eq:base_loss}, namely the Chamfer distance from predicted points to ground-truth points, the Chamfer distance from ground-truth points to predicted points, and voxel occupancy accuracy. Classes that caused larger degradation in these metrics when masked were treated as more important during pretraining.

\paragraph{Importance-Based Masking Policy}
Based on the class-importance analysis above, we used the resulting ranking to determine how the fixed LiDAR masking ratio should be distributed across semantic groups during pretraining. We first grouped all non-empty LiDAR voxels into four categories according to their semantic importance level: high, medium, low, and background. We then kept the same baseline masking ratio $\rho$, but redistributed the masked voxels across these groups so that more important semantic groups were protected more strongly, while less important and background groups were masked more heavily. For voxels containing multiple semantic classes, each voxel was assigned to the most important class present. Within each group, voxels were sampled uniformly at random. In this way, the original UniM$^2$AE pretraining protocol was preserved, while semantic structure was introduced into the LiDAR masking policy.

\subsection{Auxiliary Point-wise LiDAR Semantic Supervision}
\label{subsec:semantic_branch}

While the reconstruction losses in Section~\ref{subsec:baseline} train the model to recover masked LiDAR voxels, they do not directly encourage the learned LiDAR representations to predict semantic classes. To address this, we introduce an auxiliary point-wise LiDAR semantic supervision branch during pretraining, while keeping the downstream BEVFusion fine-tuning stage unchanged.

We attach the semantic supervision branch to the decoder side, so that it can supervise points from both masked and unmasked voxels involved in LiDAR reconstruction. Let $\Omega_{\mathrm{dec}}$ denote the set of LiDAR points with semantic labels that map to decoder voxels used for LiDAR reconstruction. For each point $p \in \Omega_{\mathrm{dec}}$, we gather the decoder-side voxel feature of its corresponding voxel, denoted by $\mathbf{f}^{\mathrm{dec}}_{p} \in \mathbb{R}^{128}$. Since multiple points can lie inside the same voxel and therefore share the same voxel-level feature, we concatenate this feature with a 3D local point offset $\Delta \mathbf{p} \in \mathbb{R}^{3}$, representing the point position relative to the voxel center, to form the point-wise semantic input $\mathbf{z}_{p} = \left[\mathbf{f}^{\mathrm{dec}}_{p} \, ; \, \Delta \mathbf{p}\right]$.

A lightweight multi-layer perceptron (MLP) semantic head $H_{\mathrm{sem}}$ takes $\mathbf{z}_{p}$ as input and predicts per-point semantic logits as $\hat{\mathbf{y}}_{p} = H_{\mathrm{sem}}(\mathbf{z}_{p})$, where $\hat{\mathbf{y}}_{p}$ denotes the predicted logit vector for point $p$, and $y_p$ denotes its ground-truth semantic label. The auxiliary semantic loss is defined as the average cross-entropy over valid labeled points, $\mathcal{L}_{\mathrm{sem}} = \frac{1}{|\Omega_{\mathrm{dec}}|} \sum_{p \in \Omega_{\mathrm{dec}}} \mathrm{CE}\!\left(\hat{\mathbf{y}}_{p}, y_{p}\right)$. The overall pretraining objective is then extended as $\mathcal{L}_{\mathrm{total}} = \mathcal{L}_{\mathrm{base}} + \lambda_{\mathrm{sem}} \mathcal{L}_{\mathrm{sem}}$, where $\mathcal{L}_{\mathrm{base}}$ is the baseline reconstruction loss defined in Section~\ref{subsec:baseline}, and $\lambda_{\mathrm{sem}}$ controls the contribution of the auxiliary semantic term. In this way, the decoder-side LiDAR features are trained not only for reconstruction, but also for semantic prediction.

\section{Experimental Evaluation}
\label{sec:experiments}

\subsection{Experimental Setup}
\label{subsec:setup}

We conducted our experiments on nuScenes mini, a subset of the nuScenes trainval split containing 10 scenes. nuScenes is a multimodal autonomous driving dataset with 360$^\circ$ surround sensing using six RGB cameras and LiDAR~\cite{nuscenes}. We used the nuScenes-lidarseg annotations, which provide 32 point-wise semantic labels for keyframe LiDAR points. Following the standard nuScenes 3D detection task, we evaluated downstream 3D BEV object detection on the 10 benchmark object categories and reported mAP, NDS, and the true-positive error metrics: mean Average Translation Error (mATE), Scale (mASE), Orientation (mAOE), Velocity (mAVE), and Attribute (mAAE).

\begin{table*}[t]
\centering
\caption{Semantic class importance analysis used to construct the Semantics-Guided LiDAR Voxel Masking policy in Section~\ref{subsec:semantic_masking}.}
\label{tab:masking_analysis}
\scriptsize
\setlength{\tabcolsep}{4pt}
\begin{tabular}{|l|c|c|c|c|c|c|}
\hline
\textbf{Detection class} & \textbf{Chamfer distance} & \textbf{Chamfer distance} & \textbf{Occupancy} & \textbf{Mean} & \textbf{Importance} & \textbf{Masking} \\
 & \textbf{GT to Pred} $\downarrow$ & \textbf{Pred to GT} $\downarrow$ & \textbf{accuracy} $\uparrow$ & \textbf{rank} & \textbf{level} & \textbf{weight} \\
\hline
car & 0.181647 & 0.436208 & 0.977032 & 7.5 & High & 0.75 \\
\hline
pedestrian & 0.180714 & 0.438906 & 0.977074 & 7.5 & High & 0.75 \\
\hline
construction\_vehicle & 0.180739 & 0.437317 & 0.976535 & 7.5 & High & 0.75 \\
\hline
motorcycle & 0.179613 & 0.436273 & 0.976579 & 5.0 & Medium & 0.95 \\
\hline
truck & 0.181256 & 0.434774 & 0.977436 & 5.5 & Medium & 0.95 \\
\hline
bus & 0.180607 & 0.437096 & 0.977038 & 6.0 & Medium & 0.95 \\
\hline
traffic\_cone & 0.181641 & 0.435008 & 0.976332 & 6.5 & Medium & 0.95 \\
\hline
barrier & 0.182271 & 0.434331 & 0.977185 & 6.5 & Medium & 0.95 \\
\hline
trailer & 0.178449 & 0.433590 & 0.976840 & 1.0 & Low & 1.05 \\
\hline
bicycle & 0.179253 & 0.434264 & 0.977334 & 2.0 & Low & 1.05 \\
\hline
background & -- & -- & -- & -- & Background & 1.20 \\
\hline
\end{tabular}
\end{table*}

\subsection{Results and Discussion}
\label{subsec:results_discussion}

Table~\ref{tab:main_results} reports downstream 3D BEV object detection results on the nuScenes mini validation set for four settings: the uniform random masking baseline from Section~\ref{subsec:baseline}, the semantics-guided LiDAR voxel masking strategy from Section~\ref{subsec:semantic_masking}, our decoder-side point semantic supervision method from Section~\ref{subsec:semantic_branch}, and a post-MMIM semantic supervision method used only as an ablation. mAP and NDS are reported in \%, where higher is better, while the true-positive error metrics are reported in their native units, where lower is better.

\paragraph{Random masking baseline}
The baseline follows the original UniM$^2$AE pretraining pipeline with uniform random LiDAR voxel masking at a fixed masking ratio of $\rho = 0.7$ (70\%). This setting achieved 24.72\% mAP and 31.41\% NDS, and serves as the reference point for evaluating the effect of introducing semantic information during pretraining.

\paragraph{Semantics-guided LiDAR voxel masking}
Semantics-guided LiDAR voxel masking keeps the same overall LiDAR masking ratio of 70\% but redistributes the masked voxels according to semantic class importance, using $\tau_c = 1$ for the class-specific voxel threshold. As shown in Table~\ref{tab:main_results}, this improves the baseline by +1.49\% mAP and +1.66\% NDS. It also yields lower true-positive error metrics than the baseline, with only a slight increase in mAVE. Overall, these results show that preserving semantically important regions during pretraining improves downstream detection.

Table~\ref{tab:masking_analysis} summarises the semantic class importance analysis used to construct the final masking policy. Since the downstream task is 3D BEV object detection on the 10 nuScenes detection categories, we mapped the 32 raw semantic labels to the corresponding detection classes wherever possible, while all remaining labels were treated as background. We then ranked the classes using voxel-level LiDAR reconstruction metrics, where lower Chamfer distances indicate better reconstruction and higher occupancy accuracy indicates better voxel prediction. Classes that caused larger degradation in reconstruction when masked were treated as more important and were therefore protected more strongly in the final masking policy. Based on this ranking, each class was assigned an importance level and a masking weight that controls how strongly that class is masked under the fixed overall LiDAR masking ratio.


\paragraph{Decoder-side point semantic supervision}
As shown in Table~\ref{tab:main_results}, the best decoder-side point semantic supervision setting achieved 26.11\% mAP and 34.63\% NDS with $\lambda_{\mathrm{sem}} = 0.25$, improving over the random masking baseline by +1.39\% mAP and +3.22\% NDS.  Compared with semantic masking, it achieved slightly lower mAP (26.11\% vs.\ 26.21\%) but higher NDS (34.63\% vs.\ 33.07\%), indicating stronger overall detection quality. It also achieved lower mATE, mASE, mAVE, and mAAE than both the baseline and semantic masking method, while mAOE remained slightly higher than semantic masking method. Overall, these results show that auxiliary point-wise semantic supervision improves downstream 3D BEV object detection and gives the best overall NDS among all evaluated settings.


\paragraph{Post-MMIM semantic supervision (ablation)}
As an ablation, we also evaluated a post-MMIM semantic supervision method in which the semantic decoder is attached after MMIM and before the decoder, using post-MMIM LiDAR voxel features together with local point offsets for semantic prediction. As shown in Table~\ref{tab:main_results}, the best post-MMIM setting achieved 26.09\% mAP and 32.96\% NDS with $\lambda_{\mathrm{sem}} = 0.25$, improving over the random masking baseline by +1.37\% mAP and +1.55\% NDS. Although this confirms that semantic supervision is beneficial even before the decoder, it remained below the method introduced in Section \ref{subsec:semantic_branch}, especially in NDS. This observation is intuitive: cascading the semantic and LiDAR reconstruction decoders enables supervision over both masked and visible voxels, whereas the post-MMIM branch primarily supervises only visible voxels.


\section{Conclusion}
\label{sec:conclusion}
In this paper, we presented a semantics-guided multimodal masked autoencoder pretraining framework that can improve downstream 3D BEV object detection. Building on the UniM$^2$AE baseline, we introduced semantic information only during pretraining through two separate extensions: semantics-guided LiDAR voxel masking, which preserves semantically important LiDAR regions more strongly, and auxiliary point-wise LiDAR semantic supervision, which complements reconstruction with explicit semantic guidance for LiDAR features. Experiments on the nuScenes mini validation set demonstrated that both extensions improved performance over the uniform random masking baseline. Semantics-guided LiDAR voxel masking improved performance by +1.49\% mAP and +1.66\% NDS, achieving the best mAP, while decoder-side point semantic supervision improved performance by +1.39\% mAP and +3.22\% NDS, achieving the best overall NDS and outperforming the post-MMIM method. Taken together, these results indicate that uniform random masking and reconstruction alone are not sufficient to learn the most effective multimodal representations, and that incorporating semantic information during pretraining leads to better downstream detection.

For future work, an important next step is to combine semantics-guided LiDAR voxel masking and auxiliary point-wise LiDAR semantic supervision within a single pretraining framework, to evaluate their joint effect after transfer to downstream detection, and to explore the use of predicted semantic labels from a semantic segmentation model instead of ground-truth semantic annotations to guide the masking stage. Another important direction is to introduce semantic information into the image branch through semantics-aware image masking or auxiliary semantic supervision for image features. It would also be valuable to evaluate the proposed ideas on larger datasets beyond nuScenes mini and to study their generalisation more broadly across additional datasets.


\bibliographystyle{ieeetr}
\newpage
\bibliography{bibliography}

\end{document}